\documentclass[runningheads]{llncs}
\usepackage[T1]{fontenc}
\usepackage{graphicx}
\usepackage{hyperref}
\hypersetup{
    pdfborder={0 0 0}
}
\usepackage{fancyhdr}
\usepackage{color}

\usepackage{booktabs}
\usepackage[table,xcdraw]{xcolor}
\usepackage{siunitx}

\begin{document}

\author{Summer Chambers\inst{} \and
Matthew C. Kelley\inst{}}

\authorrunning{S. Chambers \& M. C. Kelley}

\institute{George Mason University, Fairfax VA 22030, USA \newline  
\email{schamb3@gmu.edu}} 

\title{The misclassification of autistic writing as AI-generated}

\maketitle
\fancypagestyle{arxivfooter}{
    \fancyhf{}
    \renewcommand{\headrulewidth}{0pt}
    \fancyfootoffset[L]{1.5cm}
    \fancyfootoffset[R]{1.5cm}
    \addtolength{\footskip}{1cm} 
    \fancyfoot[C]{%
        \parbox{\dimexpr\textwidth+3cm\relax}{%
            \fontsize{8pt}{10pt}\selectfont\selectfont
This version of the contribution has been accepted for publication after peer review but is not the Version of Record and does not reflect post-acceptance improvements, or any corrections. The Version of Record is available online at: \url{https://doi.org/10.1007/978-3-031-98420-4_7}. Use of this Accepted Version is subject to the publisher’s Accepted Manuscript terms of use: \url{https://www.springernature.com/gp/open-research/policies/accepted-manuscript-terms}.}
}}
\thispagestyle{arxivfooter}
\pagestyle{fancy}
\fancyhf{}
\renewcommand{\headrulewidth}{0pt}
\fancyhead[LE,RO]{\thepage}

\begin{abstract}
 Recent findings suggest that detection models for artificial intelligence (AI) cannot accurately identify AI-generated text and may exhibit bias against certain minority groups. In the present study, anecdotal claims that autistic writers more often have their work flagged as AI-generated are examined empirically. A corpus of approximately 60,000 Reddit posts split into ``likely-autistic'' and ``general-Reddit'' subcorpora is used to compare the distribution of probabilities output by the OpenAI GPT-2 detection model. Differences in textual features between subcorpora are observed and compared to reported features of AI-generated text. Results showed that while less than two-percent of either subcorpus was flagged as AI-generated by the model, significantly more texts from the likely-autistic subcorpus were flagged. Connections between features of text with likely-autistic authors and AI-generated text were not straightforward. The widespread use of AI-detection models with a potential bias against autistic writers in their output prompts ethical scrutiny, and the authors recommend further critical examination of the models themselves as well as their use in academic contexts. 

\end{abstract}

\section{Introduction}
In the age of ChatGPT and other large language models (LLMs), schools and publishers have begun using tools to detect text written by some kind of Artificial Intelligence (AI). As ethics questions surround the creation and use of models like ChatGPT, fairness issues related to inaccurate or biased AI-detection tools have also become prominent in discourse \cite{ghaffary_universities_2023}. Concerns around AI-detection models have been bolstered by studies revealing their very low accuracy rates \cite{chaka_accuracy_2024,walters_effectiveness_2023,weber-wulff_testing_2023} as well as evidence of bias in AI detection against certain groups. In a 2023 survey, 10\% of teenagers in the US reported having been falsely accused of using AI to write their assignments. Disturbingly, twice as many Black teenagers as white and Latino teenagers reported being falsely accused \cite{madden_dawn_2024}. Additionally, Liang et al. \cite{liang_gpt_2023} showed empirically that AI-detection models are more likely to flag texts written by non-native English speakers. While there is evidence of this tendency, anecdotal claims that such models disproportionately flag autistic people's writing as AI-generated have not been formally investigated. In this paper, we describe an experiment probing the OpenAI GPT-2 detection model for a false positive bias against likely-autistic writers' posts on Reddit.

\subsection{AI Detection Models}
The GPT-2 detector used in this experiment was released in 2019 by OpenAI, the creators of the GPT series \cite{solaiman_release_2019}. The classifier purported to detect text generated with OpenAI's GPT-2 language model with 95\% accuracy, though later evaluations indicated much lower rates \cite{perkins_detection_2024}. Newer iterations of AI-detection models have since been released and subsequently deleted by OpenAI as accuracy issues persist and evidence of biases are revealed \cite{kirchner_new_2023}.

While many major universities in the US and UK are now cautioning their professors against using automated AI-detection tools, free and paid tools such as CopyLeaks, Turnitin, GoWinston, and GPTZero boast largely unverified accuracy rates up to 99.6\% and are still in high demand \cite{ghaffary_universities_2023}. Despite empirically supported claims that AI-detection models are ``neither accurate nor reliable'' \cite[p. 28]{weber-wulff_testing_2023}, their use is not explicitly prohibited in educational or publishing contexts. In fact, a 2024 survey reported that two thirds of American teachers used them regularly \cite{dwyer_report_2024}. While newer AI-detection technology called watermarking could have the potential to be much more accurate, OpenAI has not publicly released any new tools, claiming to be wary of stigmatizing the use of AI for groups who rely on it to improve their writing, such as non-native English speakers \cite{openai_understanding_2024}. 

\subsection{Understanding AI Detector Predictions}
Unlike plagiarism detectors, current AI-detection models cannot cite evidence of the phenomenon they attempt to detect, so their outputs cannot be cross-checked empirically \cite{chandere_online_2021}. It is quite hard to know what features of AI-generated text detection models rely on since they only return a prediction (``Real'' or ``Fake'') and a probability score (often taken as a model's degree of certainty in its prediction, though this interpretation can be misleading---see \cite{guo_calibration_2017}). One can learn a lot about a model from the data it was trained on. However, with large, proprietary models, data sources and statistics about those sources are rarely available.

After Liang et al. \cite{liang_gpt_2023} brought to light evidence of a bias against non-native English speakers in several popular AI-detectors---including the OpenAI GPT-2 detector---AI and plagiarism-detection company Turnitin reported that a significant difference in false positives for non-native vs. native English-speaking authors only existed for short-form texts (under the 300-word minimum suggested by Turnitin) \cite{adamson_new_2023,chaka_accuracy_2024}. Most AI-detectors recommend a minimum number of characters or words for their input, citing poor reliability under that length.  It has also been reported that low values of perplexity and burstiness---frequently associated with AI-generated text---are common in non-native English speakers' writing \cite{liang_gpt_2023}. Similar claims regarding perplexity and burstiness have not been made for autistic people's writing, but there is anecdotal evidence that humans and AI-detectors alike may be prone to mistake autistic writing styles for AI.

\subsection{Autistic Experiences with AI-Detection}
Kling \cite{kling_prof_2023} reported on a university professor's experience of being falsely accused of using AI to write their emails. The professor believed the accusation could be attributed to being autistic and mentioned that receiving such allegations is a common experience for autistic people \cite{kling_prof_2023}. An autistic student falsely accused by her professor had similar observations, citing the "formulaic" nature of her own writing and its potential similarity to AI \cite{davalos_ai_2024}. Gegg-Harrison and Quarterman \cite{gegg-harrison_ai_2024} used a small corpus of their own writing to test the false positive rates of several different popular AI-detection tools and saw much higher false positive rates than those reported by the makers of the tools themselves. They go on to discuss their own neurodivergence as a potential factor in these results, noting that a number of neurodivergent students and writers with similar suspicions reached out to them to share stories and fears related to false accusations of AI usage in their writing.

\subsection{Exploring Autistic Writing Styles}
A surprisingly small amount of descriptive research has been conducted on sociolinguistic and stylistic differences in autistic language, despite considerable literature on diagnostic linguistic traits of Autism Spectrum Disorder (ASD) largely limited to children and spoken---as opposed to written---language. Some of these linguistic traits are mentioned in diagnostic coding for the ADOS-2 \cite{woodhouse_emma_ados-2_2021} and in the DSM-5 \cite{american_psychiatric_association_diagnostic_2013}, advising diagnosticians to look out for language that is ``stereotyped'', ``formal'', ``rigid'', ``repetitive'', and ``pedantic''. While many researchers group these linguistic traits under the umbrella of pragmatic ``deficiency'' \cite{lam_pragmatic_2014}, others approach the topic more neutrally, highlighting only the divergence from the norm or majority \cite{crompton_autistic_2020,williams_mutual_2021}. Recent research spurred by autistic and other disability self-advocacy groups explicitly frames these features as ``differences'' rather than ``deficiencies'' \cite{monk_use_2022}. The finding that autistic people without language impairment communicate as effectively and efficiently as neurotypical people do when in a peer group of other autistic people \cite{crompton_autistic_2020} bolsters the claim that autism is characterized by stylistic language differences rather than pragmatic deficiencies.

At least two recent studies have attempted to make use of differences in written language to train machine learning models to predict whether or not tweets were authored by autistic users on social media platform X (formerly Twitter) using only the texts of tweets \cite{jaiswal_using_2024,rubio-martin_enhancing_2024}. While they initially claimed that the use case for such classifier models would be assisting with early autism diagnosis, Jaiswal \& Washington released an addendum to their original paper and a letter to the editor discussing serious ethical concerns with digital phenotyping and the potential for harm involved in creating such models \cite{jaiswal_addendum_2024,jaiswal_ethics_2024}. They also acknowledge a lack of explicit consent from users who authored the data that was collected, which ultimately motivated them to delete their models and data.

\subsection{The Present Paper}
In this paper, we test the hypothesis that false positives in the outputs of the OpenAI GPT-2 detection model are more common for text written by autistic people. We run a variety of Reddit posts from autism-focused and general discussion subreddits through the detector and compare its predicted probabilities of AI-generated content between the two groups of texts.

\section{Data}
While research involving public Reddit posts does not meet the criteria for submission to the authors' Institutional Review Board, ethical concerns arise when collecting data without explicit consent from its creators, especially in marginalized communities. Though this dataset contains no user-identifying information, it will not be made publicly available given the potential for malicious use of such data, as is discussed in Jaiswal et al. \cite{jaiswal_ethics_2024}. 
See additional research \cite{adams_scraping_2024,fiesler_remember_2024} on ethical considerations for projects using ``scraped'' Reddit posts.

\subsection{Data Collection}
The goal of the data collection step was to gather text written in a similar domain by autistic and non-autistic people. It is difficult to have high certainty about the accuracy of such a division, since not all autistic people identify themselves publicly as such anywhere their writing appears. The dataset used in the current study suffers from this low degree of certainty but was chosen for the quantity of publicly available and easily categorized texts. 

We identified 13 subreddits dedicated to discussing topics related to autism on the social media platform Reddit. The descriptions and rules of each of these subreddits make it clear that their purpose is to offer autistic people a space to discuss various facets of their lives. While some subreddits ask that only autistic people or only officially-diagnosed autistic people post and comment, others welcome posts from non-autistic people when made in good faith. Still, looking through a random sample of recent posts from each subreddit, the overwhelming majority are written by people who identify themselves as autistic. As in all subreddits, posters who do not follow the rules of each designated subreddit may have their posts deleted by the subreddit moderators. This means we can be slightly more certain that posts in these subreddits are going to ``belong'' or fit the theme of the subreddit. 

\begin{figure}
\includegraphics[width=\textwidth]{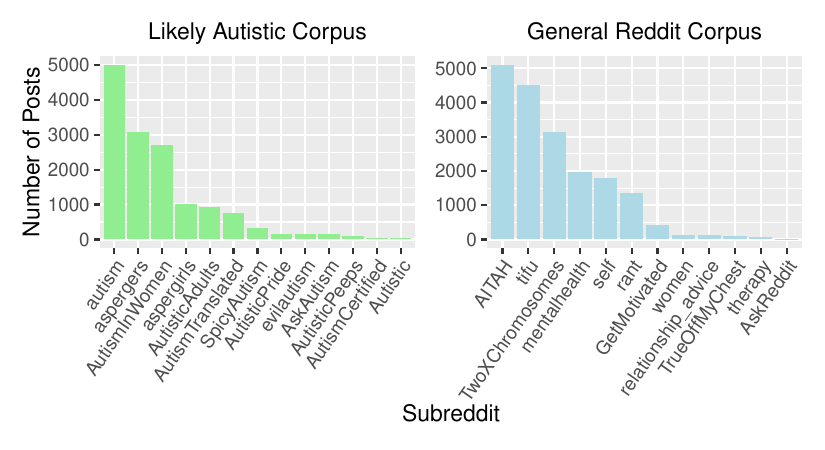}
\caption{Distribution of subreddits from which Reddit posts were collected for ``likely-autistic'' and ``general-Reddit'' subcorpora} \label{subreddits}
\end{figure}

To curate a comparison group representing the general Reddit population, we selected 12 popular subreddits judged to be similar to the autism-related subreddits in terms of post style and format. These posts tended to be relatively long first-person narratives discussing topics related to mental health, social situations, embarrassing or surprising stories, solicitations for advice or comfort, etc. Importantly, a binary distinction of ``autistic'' and ``non-autistic'' cannot be drawn between these two subcorpora, since anyone of any neurotype can join and post in any subreddit. There will be some non-autistic people posting in the subreddits targeted towards autistic people and many autistic people posting in the general subreddits. For this reason, we will refer to the two subcorpora as ``likely-autistic'' and ``general-Reddit'' hereafter. Figure~\ref{subreddits} shows the distribution of posts per subreddit present in the corpus, largely dictated by the availability of data.

We collected data with the Python \texttt{PRAW} library \cite{boe_praw-devpraw_2012}, a wrapper for the Reddit API, which only returned posts from approximately 2021-2024. With the goal of adding a larger quantity of data and less recent data, we downloaded a large set of Reddit posts from a Pushshift archive \cite{baumgartner_pushshift_2020}. This archive provided much more data than what we could get through \texttt{PRAW} alone, including posts from 2010-2020. Unfortunately, far fewer of the 13 subreddits previously identified for the autism-related subcorpus existed in the older archive. For the sake of quantity, we chose to combine the more recent and older corpora, ending up with around 60,000 Reddit posts to work with. This dataset was later halved in size, approximately, after eliminating posts with a word count of 300 or less.

Little is known about the demographics of the authors of these Reddit posts. Reddit users in these communities may be of any age, including early adolescents and adults. The overwhelming majority of texts are written in English, though varied generational, socioeconomic, cultural, and regional dialects as well as differing degrees of English proficiency are expected. All of these factors could complicate this experiment, though these issues are not easily mitigated given the anonymity of Reddit posts.

\subsection{Preprocessing and Filtering}
We used the RoBERTa AutoTokenizer which pairs with the OpenAI GPT-2 detection model for the word tokenization step. This involved truncating posts to 480 tokens, which is just below the token maximum for inputs to the OpenAI model. We also used the \texttt{NLTK} punkt sentence tokenizer \cite{loper_nltk_2002} to break posts up into sentences, though this step was purely for collecting descriptive statistics about the posts. 

In an effort to exclude as many autistic writers from the general-Reddit group as possible, we excluded posts from the general-Reddit group whose authors appeared in the likely-autistic group or which included keywords about autism in the text. In our manual review of randomly sampled posts from each subcorpus, we noticed that several of the older posts from the archived Pushshift source made in the r/autism and r/aspergers subreddits were authored by people who did not actually identify as autistic but were discussing autistic relatives. Because of this, we chose to filter out posts containing keywords about autism paired with phrases like ``my daughter…'', ``my nephew…'', etc. in an attempt to limit the number of posts from authors who are not necessarily autistic themselves. This filtering process will not catch all cases and will unnecessarily exclude some posts made by autistic people discussing relatives.

We limited our dataset to one randomly chosen post from each user to avoid over-representing any one author. We also excluded posts with fewer than 1000 characters (200 words, roughly), since OpenAI claims this is the minimum length required for an accurate prediction. 

\subsection{Corpus Statistics}
Motivated by a desire to explore textual differences between the subcorpora and their effect on AI-detection probability, we chose to compute several descriptive statistical measures for each subcorpus. For AI-generated text, both perplexity and burstiness tend to be low \cite{chaka_accuracy_2024}. Perplexity can be loosely understood as the model's degree of ``surprise'' or difficulty predicting the next word in the text. Burstiness is a term often used to represent the non-random distribution of a particular word in text, but in the context of AI-detection, it has been described as the degree of variation in sentence lengths and structures. Perplexity was computed using the \texttt{evaluate} Python package created by HuggingFace \cite{werra_trl_2020}. Burstiness was calculated as the coefficient of variation of sentence lengths for each post, modeled after the calculation used in the \texttt{zippy} Python package \cite{torrey_thinkstzippy_2024}. In addition, we computed the average word length in characters and average sentence length in words for each post in the corpus. 

\begin{figure}
\includegraphics[width=\textwidth]{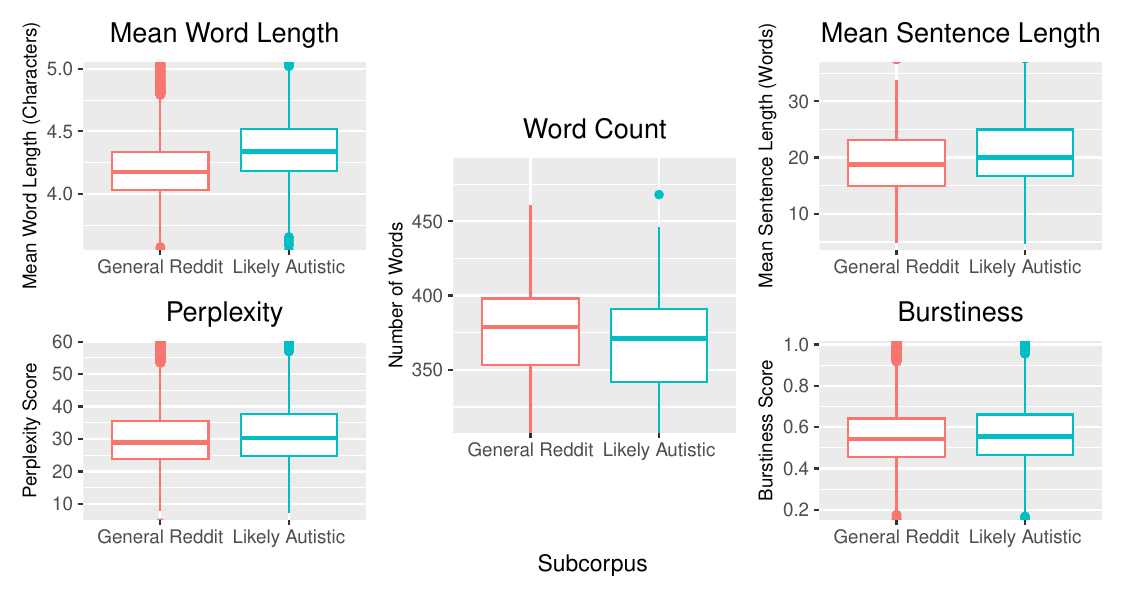}
\caption{Descriptive statistics regarding textual features for each subcorpus} \label{descriptive_stats}
\end{figure}

Most notably, the general-Reddit subcorpus had a much larger mean word count than the likely-autistic subcorpus. Mean word length and sentence length were slightly higher in the likely-autistic subcorpus, and both perplexity and burstiness were close to equal with the likely-autistic subcorpus trending only slightly higher in both. Figure~\ref{descriptive_stats} depicts the differences in textual features between the two subcorpora. 

\section{Experiment 1}

\subsection{Methods}
While there are several AI-detection tools on the market, we chose to use OpenAI's RoBERTa GPT-2 detector \cite{solaiman_release_2019} as it is freely available for download on the platform HuggingFace and thus may be more commonly used. After downloading the model locally, we ran all posts from each Reddit subcorpus through the AI-detection model to generate both binary (Fake/Real) predictions and decimal/percentage probabilities of each post being AI-generated. Predictions labeled ``Fake'' indicate an AI-probability value greater than 0.5.

There is a real risk that some of these posts actually were generated by AI, and an unequal distribution of truly AI-generated posts between the two subcorpora would compromise this experiment. One method of mitigating this risk would be to limit the data to only posts made before 2020, as GPT-2–one of the first widely available LLMs–was publicly released at the end of 2019. Still, due to the limitations of the older subcorpus, we chose to combine it with the newer subcorpus and included a binary ``date'' variable in the dataset which tracks whether or not a post was written after January 1, 2020.

In this experiment, 1.7\% of all 59,947 posts were flagged by the model as AI-generated. When split by subcorpus, the likely-autistic group had 1.9\% of posts flagged as AI-generated, and the general-Reddit group had 1.5\% flagged as such. To investigate the significance of this difference, as well as the impact of various textual features on the model's probability outputs, we fit a logistic regression model with the detection model's AI probability score as the outcome variable. The subcorpus (likely-autistic or general-Reddit) and several other variables (mean sentence length, mean word length, perplexity score, burstiness score, and a binary indication of whether the post was made after 2020) were all included as predictors. To fit this model, we used the \texttt{glm} function in \texttt{R} \cite{r_core_team_r_2024} with a binomial distribution and no random effects or interaction terms.

\subsection{Results and Discussion}

\begin{figure}
\includegraphics[width=\textwidth]{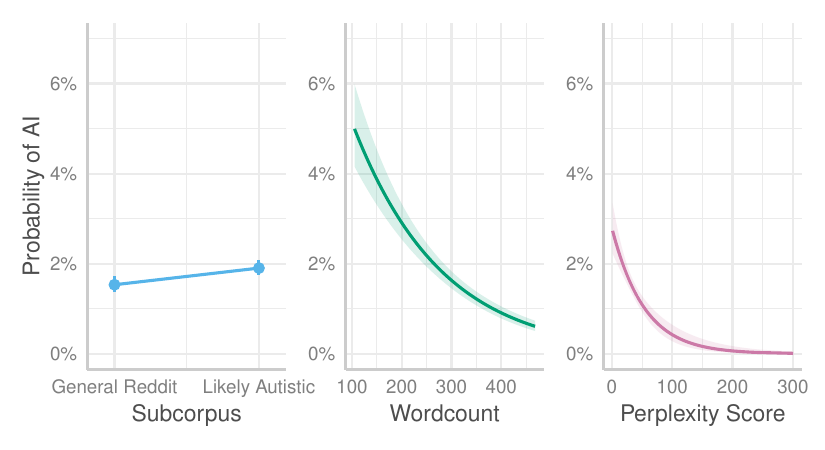}
\caption{Effect plots for significant variables in Experiment 1: \textit{Subcorpus}, \textit{Wordcount}, \textit{Perplexity Score}} \label{effects1}
\end{figure}

Table~\ref{exp1} shows the regression coefficients and Figure~\ref{effects1} shows the effect plots for significant variables of this model. We see that the subcorpus a post came from was a significant predictor in determining whether or not it was flagged as AI-generated. Based on the odds-ratio, posts from the likely-autistic subcorpus had a 25\% greater chance of being classified as AI-generated. The model also showed a significant negative effect of perplexity, which matches expectations. Posts with higher perplexity were less likely to be flagged as AI. Lastly, we see that the length of each post was significant in determining whether or not a post would be flagged as AI-generated. Shorter posts tended to be flagged as AI more often than longer posts. This tracks with the conventional wisdom that AI-detectors are less accurate with shorter texts. If the model tends to more often flag shorter texts as AI-generated, even above the 1000-character minimum threshold suggested, this is something worth exploring in greater depth.

\begin{table}
\caption{Table of coefficients for multiple logistic regression model predicting probability of text being AI-generated.}\label{exp1}
\centering
\sisetup{table-format=<2.4}
\begin{tabular}{@{}lSSSSS@{}
>{\columncolor[HTML]{FFFFFF}}l 
>{\columncolor[HTML]{FFFFFF}}l 
>{\columncolor[HTML]{FFFFFF}}l 
>{\columncolor[HTML]{FFFFFF}}l 
>{\columncolor[HTML]{FFFFFF}}l 
>{\columncolor[HTML]{FFFFFF}}l @{}}
\toprule
\multicolumn{1}{c}{\cellcolor[HTML]{FFFFFF}\textbf{Variable}} &
  \multicolumn{1}{c}{\cellcolor[HTML]{FFFFFF}\textbf{Estimate}} &
  \multicolumn{1}{c}{\cellcolor[HTML]{FFFFFF}\textbf{Std. Error}} &
  \multicolumn{1}{c}{\cellcolor[HTML]{FFFFFF}\textbf{z value}} &
  \multicolumn{1}{c}{\cellcolor[HTML]{FFFFFF}\textbf{Odds Ratio}} &
  \multicolumn{1}{c}{\cellcolor[HTML]{FFFFFF}\textbf{Pr(\textgreater{}|z|)}} \\ \midrule
Intercept            & -4.163 & 0.061 & -68.680 & 0.016 & \textless0.001 \\
Likely Autistic      & 0.220  & 0.067 & 3.305   & 1.246 & 0.001            \\
Word Count           & -0.439 & 0.031 & -14.165 & 0.644 & \textless0.001 \\
Date                 & 0.094  & 0.067 & 1.402   & 1.098 & 0.161            \\
Mean Word Length     & -0.014 & 0.027 & -0.530  & 0.986 & 0.596           \\
Mean Sentence Length & -0.016 & 0.035 & -0.457  & 0.984 & 0.648            \\
Perplexity Score     & -0.232 & 0.036 & -6.435  & 0.793 & \textless0.001 \\
Burstiness Score     & -0.012 & 0.031 & -0.406  & 0.988 & 0.685           \\ \bottomrule
\end{tabular}
\end{table}

\section{Experiment 2}

\subsection{Methods}
Given that the likely-autistic corpus had a lower mean word count than the general-Reddit corpus and that the effect of word count was significant, we decided to pursue a secondary experiment neutralizing this variable. Based on the lengths of the available data and the recommended minimum lengths from other AI-detectors, we chose to limit the corpus only to posts longer than 300 words and subsequently truncate the text of each post to exactly 300 words. In this step, the dataset size was effectively halved, resulting in a subset of $\approx$33,000 of the original $\approx$60,000 posts. The model's 500-token limit constrained our choice of maximum word count, and a larger minimum word count would have decreased the amount of data available too dramatically. Before running this set of truncated posts through the same AI-detection model as in Experiment 1, we re-computed the relevant descriptive statistics for each subcorpus for the text with normalized word counts. All other descriptive statistics showed the same comparative trends as in the previous dataset.

The following results were observed after running the 300-word posts through the same AI-detection model as in Experiment 1: 1.4\% of the 33,216 total posts were flagged as AI-generated (a slightly lower percentage than the initial experiment). 1.7\% of posts from the likely-autistic subcorpus were flagged as AI, and 1.2\% of posts from the general-Reddit subcorpus were flagged as such. We fit a new logistic regression model with this data, using the same design, method, and variables as before (with the exception of word count, which has been normalized across all posts).

\subsection{Results and Discussion}

\begin{figure}
\includegraphics[width=\textwidth]{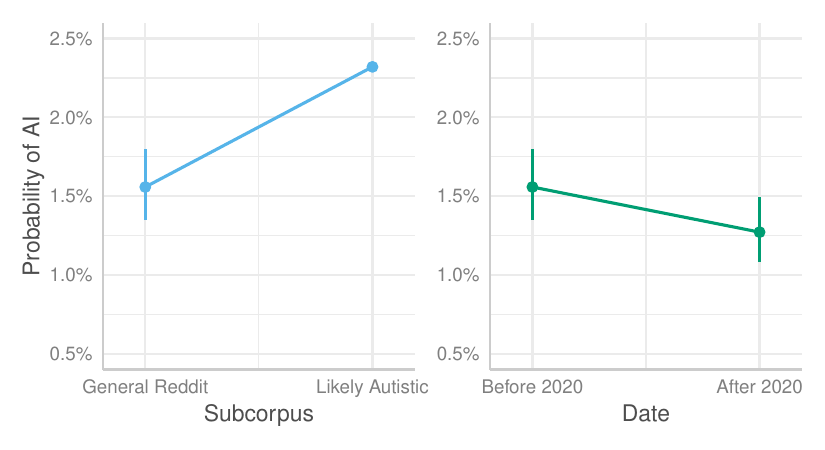}
\caption{Effect plots for significant variables in Experiment 2: \textit{Subcorpus}, \textit{Date}} \label{effects2}
\end{figure}

Table~\ref{exp2} is the table of coefficients for this regression, and effect plots for significant variables are found in Figure~\ref{effects2}. This model showed that the subcorpus variable (likely-autistic or general-Reddit) again had a significant effect on the probability of AI-generation returned by the detection model and a slightly larger effect size. Based on the odds-ratio, posts from the likely-autistic subcorpus had a 50\% greater chance of being classified as AI-generated. The ``date'' variable, a binary indication of whether or not a Reddit post was submitted after Jan 1, 2020, also showed significance in this experiment with a small effect size. The effect of this variable on AI-probability was negative, indicating that posts written in or after 2020 were less likely to be flagged as AI compared to older posts. One possible explanation for the decrease in AI-generated predictions after 2020 is that the GPT-2 detector was not trained to detect texts written by newer LLMs such as ChatGPT, so any Reddit posts written with such tools might fly under its radar as false negatives.

\begin{table}
\caption{Table of coefficients for multiple logistic regression model, Experiment 2}\label{exp2}
\centering
\sisetup{table-format=<2.4}
\begin{tabular}{@{}lSSSSS@{}
>{\columncolor[HTML]{FFFFFF}}l 
>{\columncolor[HTML]{FFFFFF}}l 
>{\columncolor[HTML]{FFFFFF}}l 
>{\columncolor[HTML]{FFFFFF}}l 
>{\columncolor[HTML]{FFFFFF}}l 
>{\columncolor[HTML]{FFFFFF}}l @{}}
\toprule
\multicolumn{1}{c}{\cellcolor[HTML]{FFFFFF}\textbf{Variable}} &
  \multicolumn{1}{c}{\cellcolor[HTML]{FFFFFF}\textbf{Estimate}} &
  \multicolumn{1}{c}{\cellcolor[HTML]{FFFFFF}\textbf{Std. Error}} &
  \multicolumn{1}{c}{\cellcolor[HTML]{FFFFFF}\textbf{z value}} 
  &
  \multicolumn{1}{c}{\cellcolor[HTML]{FFFFFF}\textbf{Odds Ratio}}
  &
  \multicolumn{1}{c}{\cellcolor[HTML]{FFFFFF}\textbf{Pr(\textgreater{}|z|)}} \\ \midrule
(Intercept)          & -4.146 & 0.075 & -55.562 & 0.016 & \textless 0.001 \\
Likely Autistic      & 0.406  & 0.096 & 4.226 & 1.501 & \textless 0.001 \\
Date                 & -0.206 & 0.095 & -2.172  & 0.814 & 0.030            \\
Mean Word Length     & -0.023 & 0.046 & -0.495  & 0.977 & 0.621                \\
Mean Sentence Length & -0.083 & 0.065 & -1.287 & 0.920 & 0.198                \\
Perplexity Score     & 0.042  & 0.043 & 0.965  & 1.043 & 0.335                \\
Burstiness Score     & 0.010  & 0.044 & 0.233  & 1.010 & 0.816               \\ \bottomrule
\end{tabular}
\end{table}

\section{General Discussion}
The fact that posts from the likely-autistic corpus were significantly more likely to be flagged as AI by this detection model prompts concern for all domains in which the model is used. False accusations of AI can cause students to suffer in terms of their academic/career standing and psychological well-being. Chaka argued that ``any AI content probability percentage or percentage point, however negligible it may be […] inflicts immeasurable reputational damage to that essay and to the student who produced it'' \cite[p. 10]{chaka_accuracy_2024}. 

Globally, autistic people suffer from extremely high rates of unemployment \cite{hong_collaborative_2024}, and reputational damage or limited educational opportunities caused by false accusations of AI will only have more devastating effects on employment rates and livelihoods. Gegg-Harrison and Quarterman \cite{gegg-harrison_ai_2024} discuss the severe psychological impact on students caused by false accusations of cheating via AI, noting that autistic people—and other neurodivergent people such as those with ADHD—often suffer from rejection sensitive dysphoria and difficulty regulating emotions, so the impacts of false accusations could be even more damaging. From another viewpoint, schools and companies could open themselves up to ableism and other discrimination lawsuits by using biased technology.

If AI-detection tools finally fall out of fashion, there is still concern that individuals will take it upon themselves to decide whether or not a text has been written by AI. It has been shown in multiple contexts and domains that humans are no better---and often worse---than automated detection models at identifying AI-generated content \cite{gao_comparing_2023}. Still, Verma \& Tenjarla \cite{verma_professor_2023} reported that Ivy League admissions officers use automated AI-detection models as well as their own judgment to decide whether or not an essay was written with AI. One admissions officer detailed a valid set of criteria he used to spot AI-generated papers which eerily echoed descriptions of autistic narrative styles (see \cite{lam_pragmatic_2014}).

There are several experimental limitations intrinsic to the data collected for this experiment. The casual style of social media text in comparison to the more formal target material of AI-detectors may constrain any generalizations made. A later iteration of this project could also put a more objective focus on matching up the subcorpora by topic. The use of other publicly available corpora will still contend with the underlying uncertainty in identifying autistic and non-autistic authors. It is reasonable to have a high degree of certainty in autistic authors' self-identifications but very difficult to know who in the ``general population'' may be autistic without identifying as such publicly. 3\% of the US population have an autism diagnosis, according to figures from 2020 \cite{maenner_prevalence_2023}. However, diagnosis rates are rising rapidly, particularly in young adults, women, and certain racial and ethnic minority groups \cite{grosvenor_autism_2024}. These rising rates seem to be an artifact of the historically widespread under-diagnosis of autism. One study reported that 80\% of autistic women were undiagnosed as of age 18 \cite{mccrossin_finding_2022}. Given these complexities, a more controlled experiment would verify all participants' results of autism evaluations when designating the two comparison groups.

The fact that we do see a significant trend of bias in the detection model's outputs even with a very noisy dataset inclines us to suspect that the difference in false positive rates between autistic and non-autistic writers may actually be larger than what was observed here. Of course, It would be decidedly more informative to get predictions from more than one AI-detection tool, as modeled in several of the previously referenced experiments. Though certain costs and inconveniences prevented the inclusion of other AI-detection tools in this study, one other freely available AI-detection tool was briefly used \cite{detector_free_2024} as an additional source of prediction data. Unfortunately, this tool produced the exact same predictions and probability scores as the OpenAI model for the same texts, rounded to the hundredth decimal place. While this is just one example, it is possible other tools on the market use the OpenAI model on the back end of their services, despite its bias and accuracy issues. 

Various seemingly incongruous descriptions of autistic language indicate that while autistic language is often more repetitive, stereotypical, and clichéd, it is also observed to be idiosyncratic and contain more neologisms or oddly-worded phrases \cite{volden_neologisms_1991}. The former point prompts the assumption that autistic writing has lower perplexity and burstiness, but in our likely-autistic subcorpus, we found perplexity and burstiness to be roughly the same as---if not slightly higher than---the general-Reddit subcorpus. Perhaps the latter observation regarding idiosyncrasy is the kernel of an explanation for why we did not see lower perplexity and burstiness. A look at the distribution of frequent lexical items and constructions in both subcorpora could give more insight into these questions.

\section{Conclusion}
Considering the prevalence of AI-generated text detection tools, it is important to understand their rates of accuracy as well as how tendencies in their outputs might affect certain groups disproportionately. This study attempted to test the hypothesis that autistic writers are more likely to have their content flagged as AI-generated by a publicly available AI-detection model. Given a large curated corpus of Reddit posts, it was shown that those posted in autism-centric subreddits---presumably written by autistic people---were more likely than posts from other subreddits to be flagged as AI-generated by OpenAI's GPT-2 detector. These findings add further motivation to examine biases in other AI-detection models and limit or discontinue their use given the potential for harm.

\begin{credits}
\subsubsection{\ackname} The authors thank James P. Blevins for his comments and thoughts on earlier versions of this project.

\subsubsection{\discintname}
The authors have no competing interests to report regarding the content of this paper.
\end{credits}

\bibliography{main}

@article{walters_effectiveness_2023,
	title = {The {Effectiveness} of {Software} {Designed} to {Detect} {AI}-{Generated} {Writing}: {A} {Comparison} of 16 {AI} {Text} {Detectors}},
	volume = {7},
	copyright = {De Gruyter expressly reserves the right to use all content for commercial text and data mining within the meaning of Section 44b of the German Copyright Act.},
	issn = {2451-1781},
	shorttitle = {The {Effectiveness} of {Software} {Designed} to {Detect} {AI}-{Generated} {Writing}},
	doi = {10.1515/opis-2022-0158},
	language = {en},
	number = {1},
	urldate = {2024-03-01},
	journal = {Open Information Science},
	author = {Walters, William H.},
	month = jan,
	year = {2023},
}

@incollection{lam_pragmatic_2014,
	address = {New York, NY},
	title = {Pragmatic {Language} in {Autism}: {An} {Overview}},
	isbn = {978-1-4614-4788-7},
	shorttitle = {Pragmatic {Language} in {Autism}},
	language = {en},
	urldate = {2024-03-03},
	booktitle = {Comprehensive {Guide} to {Autism}},
	publisher = {Springer},
	author = {Lam, Yan Grace},
	editor = {Patel, Vinood B. and Preedy, Victor R. and Martin, Colin R.},
	year = {2014},
	doi = {10.1007/978-1-4614-4788-7_25},
	pages = {533--550},
}

@article{monk_use_2022,
	title = {The use of language in autism research},
	volume = {45},
	issn = {0166-2236},
	doi = {10.1016/j.tins.2022.08.009},
	number = {11},
	urldate = {2024-03-03},
	journal = {Trends in Neurosciences},
	author = {Monk, Ruth and Whitehouse, Andrew J. O. and Waddington, Hannah},
	month = nov,
	year = {2022},
	pages = {791--793},
}

@article{crompton_autistic_2020,
	title = {Autistic peer-to-peer information transfer is highly effective},
	volume = {24},
	issn = {1362-3613},
	doi = {10.1177/1362361320919286},
	language = {en},
	number = {7},
	urldate = {2024-03-03},
	journal = {Autism},
	author = {Crompton, Catherine J and Ropar, Danielle and Evans-Williams, Claire VM and Flynn, Emma G and Fletcher-Watson, Sue},
	month = oct,
	year = {2020},
	pages = {1704--1712},
}

@article{rubio-martin_enhancing_2024,
	title = {Enhancing {ASD} detection accuracy: a combined approach of machine learning and deep learning models with natural language processing},
	volume = {12},
	issn = {2047-2501},
	shorttitle = {Enhancing {ASD} detection accuracy},
	doi = {10.1007/s13755-024-00281-y},
	number = {1},
	urldate = {2024-03-28},
	journal = {Health Inf Sci Syst},
	author = {Rubio-Martín, Sergio and García-Ordás, María Teresa and Bayón-Gutiérrez, Martín and Prieto-Fernández, Natalia and Benítez-Andrades, José Alberto},
	month = mar,
	year = {2024},
	pmid = {38455725},
	pmcid = {PMC10917721},
	pages = {20},
}

@article{chandere_online_2021,
	title = {Online {Plagiarism} {Detection} {Tools} in the {Digital} {Age}: {A} {Review}},
	copyright = {Copyright (c) 2021},
	shorttitle = {Online {Plagiarism} {Detection} {Tools} in the {Digital} {Age}},
	url = {https://annalsofrscb.ro/index.php/journal/article/view/881},
	language = {en},
	urldate = {2024-05-02},
	journal = {Annals of the Romanian Society for Cell Biology},
	author = {Chandere, Vandana and Satish, S. and Lakshminarayanan, R.},
	month = mar,
	year = {2021},
	pages = {7110--7119},
}

@article{chaka_accuracy_2024,
	title = {Accuracy pecking order – {How} 30 {AI} detectors stack up in detecting generative artificial intelligence content in university {English} {L1} and {English} {L2} student essays},
	volume = {7},
	copyright = {Copyright (c) 2024 Journal of Applied Learning and Teaching},
	issn = {2591-801X},
	doi = {10.37074/jalt.2024.7.1.33},
	language = {en},
	number = {1},
	urldate = {2024-05-02},
	journal = {Journal of Applied Learning and Teaching},
	author = {Chaka, Chaka},
	month = apr,
	year = {2024},
}

@article{perkins_detection_2024,
	title = {Detection of {GPT}-4 {Generated} {Text} in {Higher} {Education}: {Combining} {Academic} {Judgement} and {Software} to {Identify} {Generative} {AI} {Tool} {Misuse}},
	volume = {22},
	issn = {1572-8544},
	shorttitle = {Detection of {GPT}-4 {Generated} {Text} in {Higher} {Education}},
	doi = {10.1007/s10805-023-09492-6},
	language = {en},
	number = {1},
	urldate = {2024-05-03},
	journal = {J Acad Ethics},
	author = {Perkins, Mike and Roe, Jasper and Postma, Darius and McGaughran, James and Hickerson, Don},
	month = mar,
	year = {2024},
	pages = {89--113},
}

@article{jaiswal_using_2024,
	title = {Using \#{ActuallyAutistic} on {Twitter} for {Precision} {Diagnosis} of {Autism} {Spectrum} {Disorder}: {Machine} {Learning} {Study}},
	volume = {8},
	issn = {2561-326X},
	shorttitle = {Using \#{ActuallyAutistic} on {Twitter} for {Precision} {Diagnosis} of {Autism} {Spectrum} {Disorder}},
	doi = {10.2196/52660},
	urldate = {2024-05-03},
	journal = {JMIR Form Res},
	author = {Jaiswal, Aditi and Washington, Peter},
	month = feb,
	year = {2024},
	pmid = {38354045},
	pmcid = {PMC10902768},
	pages = {e52660},
}

@article{volden_neologisms_1991,
	title = {Neologisms and idiosyncratic language in autistic speakers},
	volume = {21},
	issn = {0162-3257},
	doi = {10.1007/BF02284755},
	language = {eng},
	number = {2},
	journal = {J Autism Dev Disord},
	author = {Volden, J. and Lord, C.},
	month = jun,
	year = {1991},
	pmid = {1864825},
	pages = {109--130},
}

@article{gao_comparing_2023,
	title = {Comparing scientific abstracts generated by {ChatGPT} to real abstracts with detectors and blinded human reviewers},
	volume = {6},
	issn = {2398-6352},
	doi = {10.1038/s41746-023-00819-6},
	language = {eng},
	number = {1},
	journal = {NPJ Digit Med},
	author = {Gao, Catherine A. and Howard, Frederick M. and Markov, Nikolay S. and Dyer, Emma C. and Ramesh, Siddhi and Luo, Yuan and Pearson, Alexander T.},
	month = apr,
	year = {2023},
	pmid = {37100871},
	pmcid = {PMC10133283},
	pages = {75},
}

@article{williams_mutual_2021,
	title = {Mutual ({Mis})understanding: {Reframing} {Autistic} {Pragmatic} “{Impairments}” {Using} {Relevance} {Theory}},
	volume = {12},
	issn = {1664-1078},
	shorttitle = {Mutual ({Mis})understanding},
	doi = {10.3389/fpsyg.2021.616664},
	language = {English},
	urldate = {2024-10-24},
	journal = {Front. Psychol.},
	author = {Williams, Gemma L. and Wharton, Tim and Jagoe, Caroline},
	month = apr,
	year = {2021},
}

@article{jaiswal_addendum_2024,
	title = {Addendum: {Using} \#{ActuallyAutistic} on {Twitter} for {Precision} {Diagnosis} of {Autism} {Spectrum} {Disorder}: {Machine} {Learning} {Study}},
	volume = {8},
	copyright = {Unless stated otherwise, all articles are open-access distributed under the terms of the Creative Commons Attribution License (http://creativecommons.org/licenses/by/2.0/), which permits unrestricted use, distribution, and reproduction in any medium, provided the original work ("first published in the Journal of Medical Internet Research...") is properly cited with original URL and bibliographic citation information. The complete bibliographic information, a link to the original publication on http://www.jmir.org/, as well as this copyright and license information must be included.},
	shorttitle = {Addendum},
	doi = {10.2196/59349},
	language = {EN},
	number = {1},
	urldate = {2024-10-29},
	journal = {JMIR Formative Research},
	author = {Jaiswal, Aditi and Shah, Aekta and Harjadi, Christopher and Windgassen, Erik and Washington, Peter},
	month = jul,
	year = {2024},
	pages = {e59349},
}

@article{jaiswal_ethics_2024,
	title = {Ethics of the {Use} of {Social} {Media} as {Training} {Data} for {AI} {Models} {Used} for {Digital} {Phenotyping}},
	volume = {8},
	copyright = {Unless stated otherwise, all articles are open-access distributed under the terms of the Creative Commons Attribution License (http://creativecommons.org/licenses/by/2.0/), which permits unrestricted use, distribution, and reproduction in any medium, provided the original work ("first published in the Journal of Medical Internet Research...") is properly cited with original URL and bibliographic citation information. The complete bibliographic information, a link to the original publication on http://www.jmir.org/, as well as this copyright and license information must be included.},
	doi = {10.2196/59794},
	language = {EN},
	number = {1},
	urldate = {2024-10-29},
	journal = {JMIR Formative Research},
	author = {Jaiswal, Aditi and Shah, Aekta and Harjadi, Christopher and Windgassen, Erik and Washington, Peter},
	month = jul,
	year = {2024},
	pages = {e59794},
}

@article{adams_scraping_2024,
	title = {‘{Scraping}’ {Reddit} posts for academic research? {Addressing} some blurred lines of consent in growing internet-based research trend during the time of {Covid}-19},
	volume = {27},
	issn = {1364-5579},
	shorttitle = {‘{Scraping}’ {Reddit} posts for academic research?},
	doi = {10.1080/13645579.2022.2111816},
	number = {1},
	urldate = {2024-10-30},
	journal = {International Journal of Social Research Methodology},
	author = {Adams, Nicholas Norman},
	month = jan,
	year = {2024},
	pages = {47--62},
}

@misc{werra_trl_2020,
	title = {{TRL}: {Transformer} {Reinforcement} {Learning}},
	url = {https://github.com/huggingface/trl},
	publisher = {GitHub},
	author = {Werra, Leandro von and Belkada, Younes and Tunstall, Lewis and Beeching, Edward and Thrush, Tristan and Lambert, Nathan and Huang, Shengyi and Rasul, Kashif and Gallouédec, Quentin},
	year = {2020},
}

@article{grosvenor_autism_2024,
	title = {Autism {Diagnosis} {Among} {US} {Children} and {Adults}, 2011-2022},
	volume = {7},
	issn = {2574-3805},
	doi = {10.1001/jamanetworkopen.2024.42218},
	language = {eng},
	number = {10},
	journal = {JAMA Netw Open},
	author = {Grosvenor, Luke P. and Croen, Lisa A. and Lynch, Frances L. and Marafino, Ben J. and Maye, Melissa and Penfold, Robert B. and Simon, Gregory E. and Ames, Jennifer L.},
	month = oct,
	year = {2024},
	pmid = {39476234},
	pmcid = {PMC11525601},
	pages = {e2442218},
}

@article{mccrossin_finding_2022,
	title = {Finding the {True} {Number} of {Females} with {Autistic} {Spectrum} {Disorder} by {Estimating} the {Biases} in {Initial} {Recognition} and {Clinical} {Diagnosis}},
	volume = {9},
	issn = {2227-9067},
	doi = {10.3390/children9020272},
	language = {eng},
	number = {2},
	journal = {Children (Basel)},
	author = {McCrossin, Robert},
	month = feb,
	year = {2022},
	pmid = {35204992},
	pmcid = {PMC8870038},
	pages = {272},
}

@article{maenner_prevalence_2023,
	title = {Prevalence and {Characteristics} of {Autism} {Spectrum} {Disorder} {Among} {Children} {Aged} 8 {Years} — {Autism} and {Developmental} {Disabilities} {Monitoring} {Network}, 11 {Sites}, {United} {States}, 2020},
	volume = {72},
	issn = {1546-07381545-8636},
	doi = {10.15585/mmwr.ss7202a1},
	language = {en-us},
	urldate = {2025-02-08},
	journal = {MMWR Surveill Summ},
	author = {Maenner, Matthew J.},
	year = {2023},
}

@book{american_psychiatric_association_diagnostic_2013,
	edition = {Fifth Edition},
	title = {Diagnostic and {Statistical} {Manual} of {Mental} {Disorders}},
	isbn = {978-0-89042-555-8 978-0-89042-557-2},
	language = {en},
	urldate = {2025-02-10},
	publisher = {American Psychiatric Association},
	author = {{American Psychiatric Association}},
	month = may,
	year = {2013},
	doi = {10.1176/appi.books.9780890425596},
}

@inproceedings{loper_nltk_2002,
	address = {Philadelphia, Pennsylvania, USA},
	title = {{NLTK}: {The} {Natural} {Language} {Toolkit}},
	doi = {10.3115/1118108.1118117},
	booktitle = {Proceedings of the {ACL}-02 {Workshop} on {Effective} {Tools} and {Methodologies} for {Teaching} {Natural} {Language} {Processing} and {Computational} {Linguistics}},
	publisher = {Association for Computational Linguistics},
	author = {Loper, Edward and Bird, Steven},
	month = jul,
	year = {2002},
	pages = {63--70},
}

@misc{madden_dawn_2024,
	title = {The dawn of the {AI} era: {Teens}, parents, and the adoption of generative {AI} at home and school},
	publisher = {Common Sense},
	author = {Madden, M. and Calvin, A. and Hasse, A. and Lenhart, A.},
	year = {2024},
}

@article{fiesler_remember_2024,
	title = {Remember the {Human}: {A} {Systematic} {Review} of {Ethical} {Considerations} in {Reddit} {Research}},
	volume = {8},
	issn = {2573-0142},
	shorttitle = {Remember the {Human}},
	doi = {10.1145/3633070},
	language = {en},
	number = {GROUP},
	urldate = {2025-04-24},
	journal = {Proc. ACM Hum.-Comput. Interact.},
	author = {Fiesler, Casey and Zimmer, Michael and Proferes, Nicholas and Gilbert, Sarah and Jones, Naiyan},
	month = feb,
	year = {2024},
	pages = {1--33},
}

@misc{woodhouse_emma_ados-2_2021,
	title = {{ADOS}-2 {Reliability}},
	url = {https://compasspsy.co.uk/reliability-training/ados-2/},
	language = {en-GB},
	urldate = {2024-02-04},
	author = {{Woodhouse, Emma}},
	month = aug,
	year = {2021},
}

@misc{kling_prof_2023,
	title = {Prof accused of being {AI} bot},
	url = {https://www.purdueexponent.org/campus/article_2d1826e2-2bfa-11ee-84c9-6f34496edb29.html},
	language = {en},
	urldate = {2024-02-04},
	author = {Kling, James},
	month = jul,
	year = {2023},
}

@article{verma_professor_2023,
	title = {A professor accused his class of using {ChatGPT}, putting diplomas in jeopardy},
	issn = {0190-8286},
	url = {https://www.washingtonpost.com/technology/2023/05/18/texas-professor-threatened-fail-class-chatgpt-cheating/},
	language = {en-US},
	urldate = {2024-05-01},
	journal = {Washington Post},
	author = {Verma, Pranshu},
	month = may,
	year = {2023},
}

@misc{kirchner_new_2023,
	title = {New {AI} classifier for indicating {AI}-written text},
	url = {https://openai.com/index/new-ai-classifier-for-indicating-ai-written-text},
	language = {en-US},
	urldate = {2024-05-02},
	author = {Kirchner, Jan Hendrik and Ahmad, Lama and Aaronson, Scott and Leike, Jan},
	month = jan,
	year = {2023},
}

@article{ghaffary_universities_2023,
	title = {Universities {Rethink} {Using} {AI} {Writing} {Detectors} to {Vet} {Students}’ {Work}},
	url = {https://www.bloomberg.com/news/newsletters/2023-09-21/universities-rethink-using-ai-writing-detectors-to-vet-students-work},
	language = {en},
	urldate = {2024-05-03},
	journal = {Bloomberg.com},
	author = {Ghaffary, Shirin},
	month = sep,
	year = {2023},
}

@misc{solaiman_release_2019,
	title = {Release {Strategies} and the {Social} {Impacts} of {Language} {Models}},
	urldate = {2024-10-21},
	publisher = {arXiv},
	author = {Solaiman, Irene and Brundage, Miles and Clark, Jack and Askell, Amanda and Herbert-Voss, Ariel and Wu, Jeff and Radford, Alec and Krueger, Gretchen and Kim, Jong Wook and Kreps, Sarah and McCain, Miles and Newhouse, Alex and Blazakis, Jason and McGuffie, Kris and Wang, Jasmine},
	month = nov,
	year = {2019},
	doi = {10.48550/arXiv.1908.09203},
}

@misc{openai_understanding_2024,
	title = {Understanding the source of what we see and hear online},
	url = {https://openai.com/index/understanding-the-source-of-what-we-see-and-hear-online/},
	language = {en-US},
	urldate = {2024-10-28},
	author = {{OpenAI}},
	month = aug,
	year = {2024},
}

@misc{torrey_thinkstzippy_2024,
	title = {thinkst/zippy},
	copyright = {MIT},
	url = {https://github.com/thinkst/zippy},
	urldate = {2024-10-30},
	publisher = {Thinkst Applied Research},
	author = {Torrey, Jacob},
	month = oct,
	year = {2024},
}

@misc{boe_praw-devpraw_2012,
	title = {praw-dev/praw: {PRAW}, an acronym for "{Python} {Reddit} {API} {Wrapper}", is a python package that allows for simple access to {Reddit}'s {API}.},
	url = {https://github.com/praw-dev/praw},
	urldate = {2024-10-30},
	author = {Boe, Bryce},
	year = {2012},
}

@misc{detector_free_2024,
	title = {Free {AI} {Content} {Detector} : {Tool} {To} {Detect} {Content} {Written} {By} {Ai} or {Humans}},
	shorttitle = {Free {AI} {Content} {Detector}},
	url = {https://www.freedetector.ai/},
	language = {English},
	urldate = {2024-10-30},
	author = {Detector, Free AI},
	year = {2024},
}

@misc{hong_collaborative_2024,
	title = {Collaborative {Design} for {Job}-{Seekers} with {Autism}: {A} {Conceptual} {Framework} for {Future} {Research}},
	shorttitle = {Collaborative {Design} for {Job}-{Seekers} with {Autism}},
	urldate = {2024-11-01},
	publisher = {arXiv},
	author = {Hong, Sungsoo Ray and Zampieri, Marcos and Hand, Brittany N. and Motti, Vivian and Chung, Dongjun and Uzuner, Ozlem},
	month = jul,
	year = {2024},
	doi = {10.48550/arXiv.2405.06078},
}

@misc{adamson_new_2023,
	title = {New research: {Turnitin}'s {AI} detector shows no statistically significant bias against {English} {Language} {Learners}},
	url = {https://www.turnitin.com/blog/new-research-turnitin-s-ai-detector-shows-no-statistically-significant-bias-against-english-language-learners},
	language = {en-us},
	urldate = {2025-01-26},
	author = {Adamson, David},
	month = oct,
	year = {2023},
}

@book{r_core_team_r_2024,
	address = {Vienna, Austria},
	title = {R: {A} {Language} and {Environment} for {Statistical} {Computing}},
	url = {https://www.R-project.org/},
	publisher = {R Foundation for Statistical Computing},
	author = {{R Core Team}},
	year = {2024},
}

@article{davalos_ai_2024,
	title = {{AI} {Detectors} {Falsely} {Accuse} {Students} of {Cheating}—{With} {Big} {Consequences}},
	url = {https://www.bloomberg.com/news/features/2024-10-18/do-ai-detectors-work-students-face-false-cheating-accusations},
	language = {en},
	urldate = {2025-02-10},
	journal = {Bloomberg.com},
	author = {Davalos, Jackie and Yin, Leon},
	month = oct,
	year = {2024},
}

@misc{dwyer_report_2024,
	title = {Report – {Up} in the {Air}: {Educators} {Juggling} the {Potential} of {Generative} {AI} with {Detection}, {Discipline}, and {Distrust}},
	shorttitle = {Report – {Up} in the {Air}},
	url = {https://cdt.org/insights/report-up-in-the-air-educators-juggling-the-potential-of-generative-ai-with-detection-discipline-and-distrust/},
	language = {en-US},
	urldate = {2025-02-10},
	author = {Dwyer, Maddy and Laird, Elizabeth},
	month = mar,
	year = {2024},
}

@incollection{gegg-harrison_ai_2024,
	title = {{AI} {Detection}'s {High} {False} {Positive} {Rates} and the {Psychological} and {Material} {Impacts} on {Students}},
	copyright = {Access limited to members},
	isbn = {9798369302408},
	language = {en},
	urldate = {2025-05-02},
	booktitle = {Academic {Integrity} in the {Age} of {Artificial} {Intelligence}},
	publisher = {IGI Global Scientific Publishing},
	author = {Gegg-Harrison, Whitney and Quarterman, Claire},
	year = {2024},
	doi = {10.4018/979-8-3693-0240-8.ch011},
	pages = {199--219},
}

@inproceedings{guo_calibration_2017,
	address = {Sydney, NSW, Australia},
	series = {{ICML}'17},
	title = {On calibration of modern neural networks},
	doi = {10.48550/arXiv.1706.04599},
	urldate = {2025-05-01},
	booktitle = {Proceedings of the 34th {International} {Conference} on {Machine} {Learning} - {Volume} 70},
	publisher = {JMLR.org},
	author = {Guo, Chuan and Pleiss, Geoff and Sun, Yu and Weinberger, Kilian Q.},
	month = aug,
	year = {2017},
	pages = {1321--1330},
}

@article{baumgartner_pushshift_2020,
	title = {The {Pushshift} {Reddit} {Dataset}},
	volume = {14},
	copyright = {Copyright (c) 2020 Association for the Advancement of Artificial Intelligence},
	issn = {2334-0770},
	doi = {10.1609/icwsm.v14i1.7347},
	language = {en},
	urldate = {2025-05-02},
	journal = {Proceedings of the International AAAI Conference on Web and Social Media},
	author = {Baumgartner, Jason and Zannettou, Savvas and Keegan, Brian and Squire, Megan and Blackburn, Jeremy},
	month = may,
	year = {2020},
	pages = {830--839},
}

@article{liang_gpt_2023,
	title = {{GPT} detectors are biased against non-native {English} writers},
	volume = {4},
	issn = {2666-3899},
	doi = {10.1016/j.patter.2023.100779},
	number = {7},
	urldate = {2025-05-02},
	journal = {Patterns},
	author = {Liang, Weixin and Yuksekgonul, Mert and Mao, Yining and Wu, Eric and Zou, James},
	month = jul,
	year = {2023},
	pages = {100779},
}

@article{weber-wulff_testing_2023,
	title = {Testing of detection tools for {AI}-generated text},
	volume = {19},
	copyright = {2023 The Author(s)},
	issn = {1833-2595},
	doi = {10.1007/s40979-023-00146-z},
	language = {en},
	number = {1},
	urldate = {2025-05-02},
	journal = {Int J Educ Integr},
	author = {Weber-Wulff, Debora and Anohina-Naumeca, Alla and Bjelobaba, Sonja and Foltýnek, Tomáš and Guerrero-Dib, Jean and Popoola, Olumide and Šigut, Petr and Waddington, Lorna},
	month = dec,
	year = {2023},
	pages = {1--39},
}
\end{document}